# Deep Neural Network Based Real-time Kiwi Fruit Flower Detection in an Orchard Environment

[1]JongYoon Lim, [2]Ho Seok Ahn, [3]Mahla Nejati, [4]Jamie Bell, [5]Henry Williams and [6]Bruce A. MacDonald
Department of Electrical, Computer and Software Engineering, CARES, University of Auckland
Auckland, New Zealand
[1, 2, 5, 6]{jy.lim, hs.ahn, henry.williams, b.macdonald}@auckland.ac.nz, [3, 4]{mnej691, jlee611}@aucklanduni.ac.nz

## Abstract

In this paper, we present a novel approach to kiwi fruit flower detection using Deep Neural Networks (DNNs) to build an accurate, fast, and robust autonomous pollination robot system. Recent work in deep neural networks has shown outstanding performance on object detection tasks in many areas. Inspired this, we aim for exploiting DNNs for kiwi fruit flower detection and present intensive experiments and their analysis on two state-of-the-art object detectors; Faster R-CNN and Single Shot Detector (SSD) Net, and feature extractors; Inception Net V2 and NAS Net with real-world orchard datasets. We also compare those approaches to find an optimal model which is suitable for a real-time agricultural pollination robot system in terms of accuracy and processing speed. We perform experiments with dataset collected from different seasons and locations (spatio-temporal consistency) in order to demonstrate the performance of the generalized model. The proposed system demonstrates promising results of 0.919, 0.874, and 0.889 for precision, recall, and F1-score respectively on our real-world dataset, and the performance satisfies the requirement for deploying the system onto an autonomous pollination robotics system.

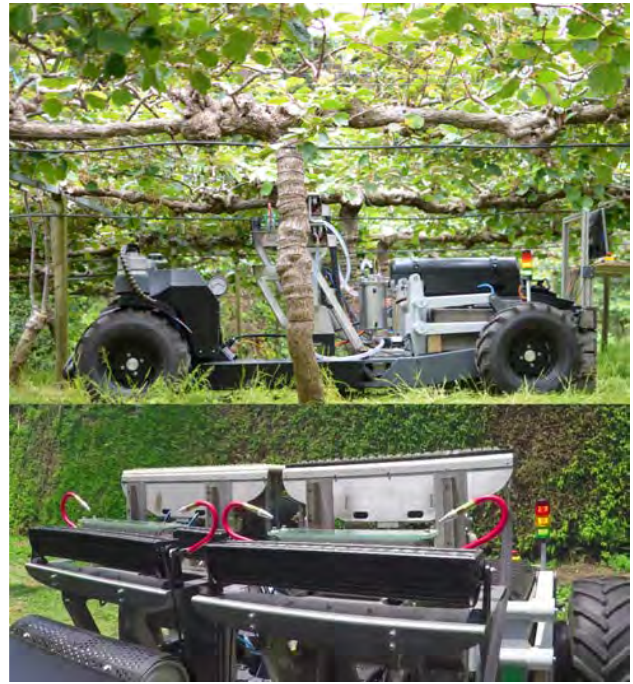

Fig. 1. The pollination robot system designed to individually track and pollinate kiwi fruit flowers with a liquid pollen solution.

## 1 Introduction

We are using lots of autonomous robots in different areas, and there are notable differences between them. Robots in orchard environments should consider challenges such as the presence of uncontrolled growth, exposure to the weather as well as the slope, softness of the ground, cluttered and undulating nature of terrain in orchards, although there are also many similarities in the technologies used across different robots, both indoors and outdoors [Bell et al., 2016]. In particular, the outdoor orchard environment gives fickle conditions that lead to bog challenges for developing an autonomous system [Li et al., 2009].

Outdoor pollination is one of the challenges to us as bees are dying at alarming rates due to colony collapse disorder, pesticides, and invasive mites. Some studies show that between 2015 and 2016 beekeepers lost 44% of bee colonies over the winter [Aizen et al., 2009]. Additionally, caring for bees is another challenge to orchard owners. Even caring for human labour is a challenge as it is difficult to hire seasonal workers.

To solve these issues, rather than relying on bees and manual pollination by human workers, researchers develop pollination robots that spray pollen on flowers. Another benefit of an accurate pollination robot is that it contributes to keeping the fruits at a uniform size, quality, and ripening to make them more valuable products. Other applications of spraying targeted at flowers include spraying a reagent to precisely control uniform fruit ripening time [Kurosaki et al., 2012]. Ting Yuan et al. spray hormones on tomato flowers for the same reason [Yuan et al., 2016].

Accurate flower detection is one of the critical technologies for successfully targeted flower spraying, such as pollination. Many earlier flower detection methods use color values as target features, i.e., [Das et al., 1999], but it did not work well due to similar colors of flowers. Therefore, researchers started to include more information on the detection process such as size, shape, and/or edge of features. Maria-Elena Nilsback et al. used color and shape [Nilsback and Zisserman, 2006]. Graph cut and color-dependent conditional random forest algorithms were used to obtain color features.

For shape features, generic shape fitting was used to detect petals. Chomtip Pornpanomchai et al. used Red, Green and Blue values with the flower size and the edge of the petals feature to find herb flowers [Pornpanomchai et al., 2017]. Soon-Won Hong et al. detects the contour of a flower image by using both color-based and edge-based contour detection [Hong and Choi, 2012]. Then, we classify its color groups and contour shapes by using k-means clustering and history matching. A. Abinaya et al. detected Jasmine flower using the thresholding technique as well [Abinaya and Roomi, 2016]. However, these color-based approaches are not robust enough on different lighting conditions.

Recently researchers use machine learning and Deep Neural Network (DNN) techniques for flower detection [Ahn et al., 2018]. Yahata et al. proposed image sensing methods that are constructed by combining several image processing and machine learning techniques [Yahata et al., 2017]. The flower detection is realized based on a coarse-to-fine approach where candidate areas of flowers are firstly detected by Simple Linear Iterative Clustering (SLIC) and hue information, and the acceptance of flowers is decided by Convolutional Neural Network (CNN). Yuanyuan Liu et al. also developed a flower detection method based on CNN [Liu et al., 2016]. Inkyu et al. proposed a fast and reliable fruit detection system using deep convolutional neural networks with the imagery obtained from two modalities: colour (RGB) and Near-Infrared (NIR) [Sa et al., 2016]. Pratul P. Srinivasan et al. developed a machine learning algorithm that takes as input a 2D RGB image and synthesizes a 4D RGBD light field (color and depth of the scene in each ray direction) [Srinivasan et al., 2017]. It consists of a CNN that estimates scene geometry, a stage that renders a Lambertian light field using that geometry, and a second CNN that predicts occluded rays and non-Lambertian effects. These approaches show better results than the traditional methods, but this requires appropriate training modeling with a large dataset.

In this paper, we propose a DNN based kiwi fruit flower detection method. Kiwi fruit orchards are a pergola environment [Bell et al., 2016], which means flowers are hanging down from the canopy and mainly facing to the ground, but all are facing different directions. While many kiwi fruit flowers can be seen with the cameras facing upward, this also leads to other challenges such as sunshine, which makes challenging illumination conditions. In this environment, it is quite difficult to use the earlier flower detection methods, which use color values. Therefore, we design different combinations of DNN meta-architectures and feature extractors and test them to find the best kiwi fruit flower detection method. Fig. 1 shows the orchard robot system using our kiwi fruit flower detection module for kiwi fruit flower pollination [Williams et al., 2019a; Williams et al., 2019b; Williams et al., 2019c].

The remainder of this paper is structured as follows. Section II introduces our meta-architectures and feature extractors, and Section III describes our training methods for kiwi fruit flower detection, such as data collection, labeling, and training. We present our experimental system in Section IV, and show the experimental results in Section V. We conclude this paper in Section VI.

## 2 DNN Based Detection Method

Neural nets have become the leading method for high-quality object detection in recent years [Huang et al., 2017] due to the use of CNNs. Recently developed object detectors based on these networks, such as Faster R-CNN [He et al., 2016], R-FCN [Dai et al., 2016], and SSD [Lui et al., 2015], are considered to be achieving human-level performance in terms of accuracy [Geirhos et al., 2017]. Some of them are fast enough to be run on mobile devices. These methods offer flexibility, robustness, and faster inference time.

In our paper, we focus on SSD and Faster R-CNN meta-architectures. As shown in Fig. 2, Faster R-CNN models are better suited to cases where high accuracy is desired, and latency is of lower priority. Conversely, if processing time is the most crucial factor, SSD models are recommended [Huang et al., 2017]. Speed and accuracy depend on many other factors, such as which feature extractor is used, input image sizes, etc. However, we try to compare which architecture is more suitable for our system in terms of execution time and accuracy.

### 2.1 Meta-Architecture for object detection

#### 2.1.1 Faster R-CNN

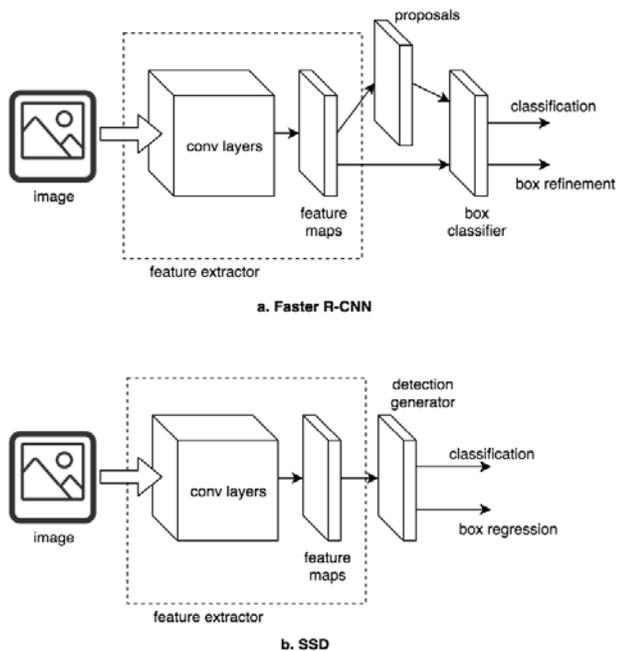

Fig. 2. Meta-Architectures which were used for our experiment. Faster R-CNN uses two step processing which leads better performance than SSD in terms of accuracy. On the other hand, SSD has faster inferencing than Faster R-CNN.

Faster R-CNN is a meta-architecture that generates box proposals using neural networks, as shown in Fig. 2.a. There exist similar variants that are relied on an external proposal generator (e.g., R-CNN or Fast R-CNN). Faster R-CNN additionally introduced the Region Proposal Network (RPN), which shares convolutional features with the classification network, and two networks are concatenated as one network that can be trained and tested through an end-to-end process.

Faster R-CNN consists of two parts: a region proposal network (RPN) and a region classifier. In the first stage, images are processed by a feature extractor, such as Inception V2 [Szegedy et al., 2015] or VGG16 [Simonyan and Zisserman, 2014], and features at some selected intermediate layer are used to predict class agnostic box proposals. In the second stage, these box proposals are used to crop features from the same intermediate feature map which are subsequently fed to the remainder of the feature extractor to predict a class and class-specific box refinement for each proposal.

*2.1.2 SSD*

SSD, shown in Fig. 2.b, is a meta-architecture that uses a single feed-forward convolutional network to directly predict class probabilities and anchor offsets without requiring a proposal generation and subsequent feature resampling stages. SSD discretizes the output space of bounding boxes into a set of default boxes over different aspect ratios and scales per feature map location. At prediction time, the network generates scores for the presence of each object category in each default box and produces adjustments to the box to better match the object shape. Additionally, the network combines predictions from multiple feature maps with different resolutions to naturally handle objects of various sizes.

**2.2 Feature Extractor for Object Detection**

Before carry out object detection tasks, it is essential to extract distinguishable features from the input images. Exploiting convolutional feature extractors which encapsulate not only low-level features such as edges, textures, but high-level semantic information such as shape, or geometric relationship is one of the convincing approaches. In this section, we present two popular feature extractors used in this paper; Inception v2 and NAS [Zoph et al., 2017]. Fig. 3 shows the basic concept of Inception and NAS.

*2.2.1 Inception V2 Network*

The main idea of the Inception architecture is based on finding out how an optimal local sparse structure in a convolutional vision network can be approximated and covered by readily available dense components. To achieve this (see Fig. 3.a), it performs convolutions with three different sizes of filters (1x1, 3x3, 5x5) on an input data. Additionally, max pooling on input data is performed, and all the outputs are concatenated for the input of the next Inception module. It also constrains the number of input channels by adding an extra 1x1 convolution before the 3x3 and 5x5 convolutions in order for reducing computational loads.

One of the advantages of the Inception network is that it includes a variety of parameter combinations (e.g., varying kernel sizes, or max pooling) into its architecture, whereas other networks utilize fixed parameters, which is one of the significant drawbacks of DNN. This, in turn, lets the network learn better feature representations and is suitable for our kiwi fruit flower detection.

*2.2.2 NAS Network*

The main idea of NAS is to design a novel search space, such that the best architecture learned on the CIFAR-10 dataset would be transferred to image classifications across a range of data and computational scales. In NAS, a controller Recurrent Neural Network (RNN) samples child networks with different architectures. The child networks are trained to convergence to obtain some accuracy on a held-out validation set. The resulting accuracies are used to update the controller so that the controller can generate better architectures over time.

**3 Training Model**

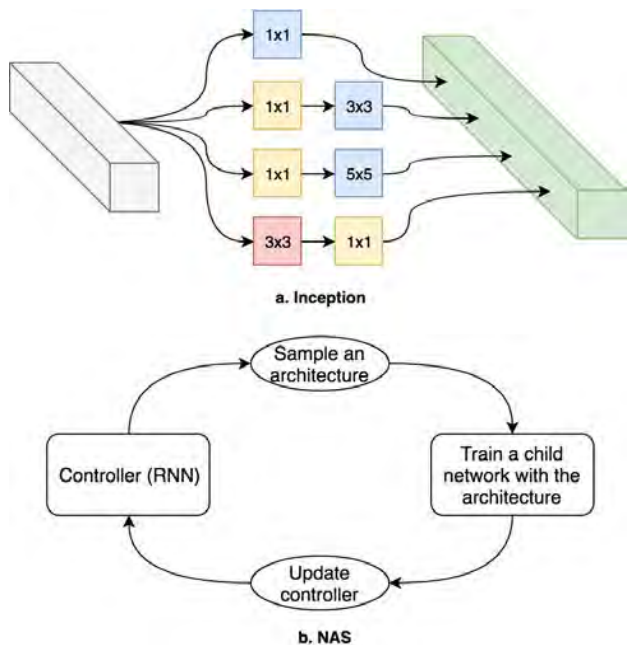

Fig. 3. One of main concepts of Inception and NAS network architecture.

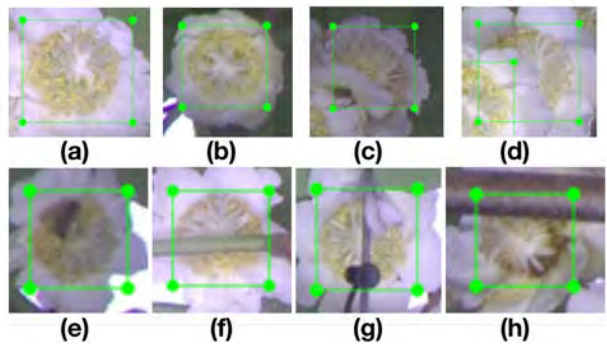

Fig. 4. Examples of Bounding box for stigma. We labeled (a) open, and (b) half opened flowers. Also occluded flowers by (c-d) flower, (e) bee, and (f-h) branch and wire were labeled.

## 3.1 Data collection

We collected green kiwi fruit flower dataset from two different kiwi orchards using a calibrated stereo camera system mounted on a mobile robot platform. The robot traversed inter-rows in the orchards while taking high fidelity images (1920x1080) at 20Hz. The data collection campaigns spontaneously performed over fruit flower season in order to capture temporal changes in flower growth. In total, 1451 images were labeled, and we use 1015 for the model training and 436 for testing that will be explained in the following sections.

## 3.2 Data labeling

For the supervised object detection task, we are required to create bounding boxes and the corresponding labels for stigmas. We use a labeling tool named Labelbox (https://labelbox.com) and have one class (i.e., kiwi fruit flower). It is worth mentioning that we only label stigmas of flower instead of the whole flower with petal because the stigmas are our target for pollination and we can detect the center of the flower in case flowers are clustered (see Fig. 4). Also, we label all stigmas as the same class regardless of male or female flowers (distinguishing these is a different problem, inter-class object detection, and is out of the scope of this paper).

We draw the bounding boxes around the stigmas aligning the centre of the boxes with the centre of the stigma. The annotated dataset includes both fully opened flowers, and half-opened ones because the latter could also get sprayed for pollination. We have intensively annotated all the flowers that humans could detect in a given image, including tilted flowers, occluded stigmas due to bees or branch and one with varying illumination. We experienced blurry images due to pollination spray drops and no petal flowers, which also included in our dataset since these cases could happen in real-world experiments. To obtain consistent labeling results, we follow clear annotation guidelines describing what we have mentioned. Fig. 4 shows examples of labeled stigma with bounding box.

## 3.3 Training Network

We use the TensorFlow (https://www.tensorflow.org) Object Detection API for training kiwi fruit flower detection models. The TensorFlow Object Detection API is an open-source framework built on top of TensorFlow that makes it easy to construct, train, and deploy object detection models.

Training a state-of-the-art object detector from scratch can take days on datasets such as COCO [Lin et al., 2014], across multiple GPUs. In order to speed up training, we take an object detector trained on different objects (COCO) and reuse the feature extractor parameters from a pre-existing object classification or detection checkpoint to initialize our new model. We try several combinations of hyper-parameters of the training configuration. We focus on finding the best combination of the input image size and training batch size on our training hardware environment.

On the other hand, different combinations of the number of regional proposals and data augmentation options do not have a meaningful influence on our experiment of kiwi fruit flower detection. We use default values for other hyper-parameters which were used for the pre-trained model (trained on the COCO dataset). We choose 1015 images and 436 images for training and validation, respectively. We stop training when accuracy and loss look saturated enough, and it usually takes 1-2 days for that on our training environment.

## 4 Experimental System

### 4.1 Overview

We create a vision processing system for detecting objects from the stereo images input to generate 3-dimensional positions for each object. The vision processing system is a part of the orchard kiwi fruit flower pollination robotic system. This package publishes the detected 3-dimension points and an action planning scheduler subscribes to these messages. The scheduler calculates when the sprayer should shoot pollination liquid at flowers based on the detected 3-dimensional points, velocity & acceleration of robot vehicle. The system is built on ROS and every module is implemented as a ROS Node. When it comes to the Vision Processing System, it consists of three nodes as shown in Fig. 5; 1) Object Detection node, 2) Stereo Matching node, and 3) 3-dimentional Position Calculation node.

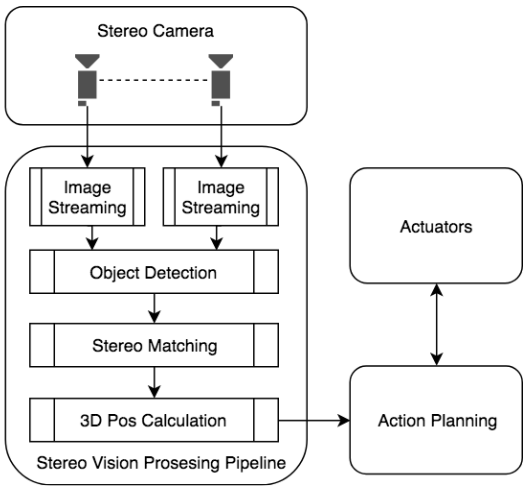

Fig. 5. System architecture of the pollination robot system.

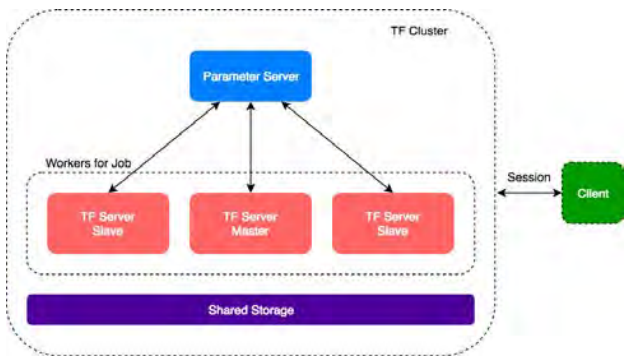

Fig. 6. Distributed TensorFlow training system. This distributed system is 1.8x faster than single TensorFlow machine on training.

### 4.2 Training System Environment

For the training of DNN-based object detection model, we use 3 high performance PCs with GPU processing and 1 standard pc. Firstly, we setup a distributed TensorFlow training system which consists of 3 GPU machines for clustered workers and 1 machine for a parameter server as shown in Fig. 6. Our system trains a model 1.8X faster than single GPU training. A machine for training has NVIDIA P6000, 32GB ram, iCore7 CPU and 500GB SSD. And a machine for parameter server has 16GB ram, iCore5 CPU and 1TB disk. We create a Docker [Merkel 2014] image and container for having the same environment across multiple machines. And a Samba server is used for sharing the dataset between working machines. Detailed hardware specification of each PC and software versions are shown in Table 1 and 2, respectively.

### 4.3 Execution System Environment

We deploy the models we train to our pollination robot system which has 4 PCs with GPU processing. Each PC has 2 GPU cards to inference Kiwi fruit flowers on stereo images from camera systems at the same time.

### 4.4 Deployment trained models

We use a version control system such as git to store models and their configurations. We decide to use branches to save each model rather than creating a repository for each. And the size of model file is usually much bigger than a normal text file, we have to use LFS (Git Large File Storage) to store them in the git server. Later we switch to google drive because of easy accessibility than git.

### 4.5 Logging System

ROS logging is used to record log messages. To log image files from the results of detection and stereo matching, we develop a library which quickly saves images into disk without memory overflow problems.

## 5 Experimental Results

### 5.1 Traditional color detection

Previously, we used a color filter method to detect kiwi fruit flowers. For this, we divided our detection pipeline into four stages; 1) Finding white and yellow region of flowers using an HSV color filter [Kim et al., 2014], 2) Using a salience map [Kim et al., 2014] by considering different resolution of images, 3) Merging the two above filters in order to detect regions of the flower, and 4) Using blob detection by setting the convexity and size of the flower in order to detect the flower separately and report them by their positions and centers. With this approach, we obtained 0.8 for precision, 0.57 for recall and 0.66 for F1 score on high quality images which were taken under stable light conditions. This detection method had almost real-time performance as 2.9ms processing time per image but it did not have robustness to external factors such as lighting changes. Therefore, that resulted in spending a long time to find best parameters for this model before testing in the outdoor environment.

|  | TF Server | Parameter Server |
|---|---|---|
| Processor | Intel Xeon 16 Core | Intel i7 4 Core |
| Memory | 32GB | 16GB |
| Graphics | Nvidia P6000 | Intel HD 4600 |
| Disk | SSD 500GB | HDD 1TB |

Table 1. Hardware specification for object detection model training.

| OS | Ubuntu 16.04 |
|---|---|
| DNN Platform, Library | Nvidia CUDA8, cuDNN5 |
| GPU Driver | Nvidia driver 384 |
| Machine Learning Framework | Tensorflow 1.4 |
| Virtual Environment | Docker ce-17 |
| Python | Python 2.7 |

Table 2. Software versions for object detection model training.

| Model Name | Speed (ms) | COCO mAP |
|---|---|---|
| Faster R-CNN NAS | 1833 | 43 |
| Faster R-CNN Inception V2 | 58 | 28 |
| SSD Inception V2 | 42 | 24 |

Table 3. Object detection models for our experiments.

| Dataset | #Avg Flower | Glare | Ripeness | #Images |
|---|---|---|---|---|
| A | 60 | Normal | Fully | 62 |
| B1 | 20 | Glare | Fully | 68 |
| B2 | 30 | Normal | Fully | 82 |
| B3 | 15 | Normal | Half | 69 |

Table 4. Details of dataset.

### 5.2 Models Comparison

In this paper, we focus on comparing DNN object detection models (not the previous color filter model). Table 3 shows a brief description of configuration for each model.

### 5.3 Dataset for Evaluation

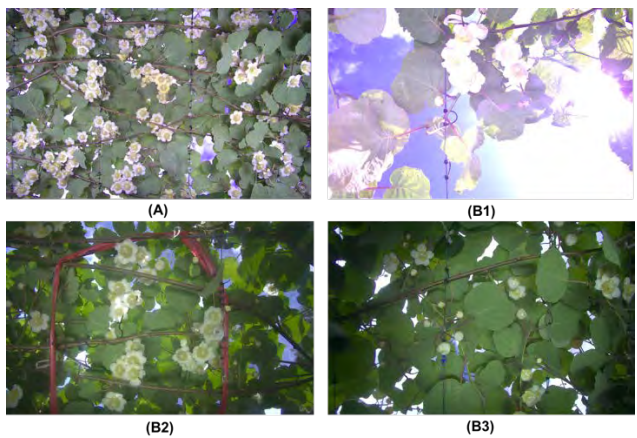

Fig. 7. Kiwi fruit flower images captured in different orchards to collect data on various conditions.

Images are captured using the same stereo camera system on the mobile robot vehicle. So, datasets for evaluation include images taken continuously by both left and right camera systems while moving at human walking speed. We choose images from different orchards for evaluation. We divide the evaluation group into A and B. Dataset A includes images taken further than Dataset B from the canopy, so it has more flowers in an image, as shown in Fig. 7. Dataset B includes images taken closer than Dataset A, so flowers look bigger than Dataset A. Also, Dataset B is divided into B1, B2, and B3 groups. Dataset B1 includes only glare images caused by sunlight because we expected it could be challenging to detect flowers from those. Dataset B2 is collected during the best time to pollinate. Dataset B3 has flower images taken just before pollination season started, so not many flowers are opened. Details of datasets for evaluation are described in Table 4. The images for evaluation are not used to train or validate our detection models.

## 5.4 Results

Evaluation is taken based on 0.5 IOU. Overall, Faster R-CNN NAS gets the highest precision score and Faster R-CNN Inception V2 has the highest Recall and F1 Score (Table 5). For Dataset A, the result shows a similar shape to the overall results, but the performance of SSD Inception V2 is much lower than the overall result (Table 6). For Dataset B, NAS has the highest score on Precision and F1 Score rate and Faster R-CNN Inception V2 has the highest score on Recall rate (Table 7). However, SSD Inception V2 has better performance than Faster R-CNN Inception V2 in terms of precision. We choose Faster R-CNN Inception V2 for further evaluation because it shows the best performance and is considered suitable for a real-time robot system thanks to the fast inference time. Results are shown in Table 5, 6, and 7, respectively.

We proceed with more experiments with the Faster R-CNN Inception V2 model, which has excellent performance on Dataset B1, B2, and B3, which include glare, normal and pre-mature flower images, respectively. It has high performance on Dataset B1 and B2, and it means this object detection model is robust against glare light condition resulted from the sun. Table 8 shows the evaluation result. Fig. 8 shows that Faster R-CNN Inception V2 has the highest rate on TP and the lowest rate on FN. And Fig. 9 shows the inference results by Faster R-CNN NAS, SSD Inception V2, and Faster R-CNN Inception V2.

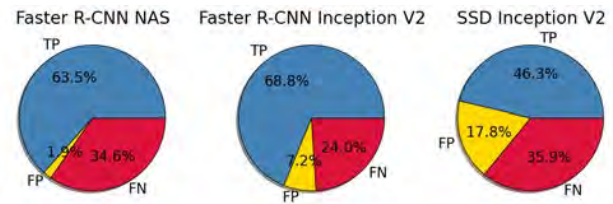

Fig. 8. Detection results using Faster R-CNN NAS, Faster R-CNN Inception V2 and SSD Inception V2 on the evaluation dataset.

| Model | Precision | Recall | F1 Score |
|---|---|---|---|
| Faster R-CNN NAS | **0.968** | 0.680 | 0.790 |
| Faster R-CNN Inception V2 | 0.904 | **0.758** | **0.820** |
| SSD Inception V2 | 0.785 | 0.612 | 0.681 |

Table 5. Results for all the Datasets.

| Model | Precision | Recall | F1 Score |
|---|---|---|---|
| Faster R-CNN NAS | **0.974** | 0.594 | 0.736 |
| Faster R-CNN Inception V2 | 0.900 | **0.721** | **0.800** |
| SSD Inception V2 | 0.632 | 0.505 | 0.560 |

Table 6. Results for Dataset A.

| Model | Precision | Recall | F1 Score |
|---|---|---|---|
| Faster R-CNN NAS | **0.962** | 0.767 | **0.843** |
| Faster R-CNN Inception V2 | 0.907 | **0.795** | 0.841 |
| SSD Inception V2 | 0.937 | 0.719 | 0.801 |

Table 7. Results for Dataset B.

| Model | Precision | Recall | F1 Score |
|---|---|---|---|
| Dataset B1 | 0.911 | **0.874** | **0.889** |
| Dataset B2 | **0.919** | 0.790 | 0.845 |
| Dataset B3 | 0.893 | 0.730 | 0.796 |

Table 8. Evaluation results of Faster R-CNN Inception V2 on Dataset B1, B2 and B3.

## 6 Conclusions

In this paper, we have trained and compared different models for kiwi fruit flower detection. Our experiments show that Faster R-CNN Inception V2 model can make fast and stable predictions about kiwi fruit flowers under the canopy in the real orchards. This model is deployed on our pollination robot system for our final evaluation.

For this work, we collect various data from two kiwi fruit orchards between early flower season and pollination season. We label kiwi fruit flowers in the dataset under strict guidelines. To find a suitable model for real-time pollination robot, we choose state-of-the-art deep neural network models for object detection and compare them from a performance perspective. We found that a transfer learning technique, where a model trained on the COCO dataset is repurposed on a task to find kiwi fruit flowers, is effective and efficient in terms of performance and training time, respectively. The trained model has fast and robust performance in the real orchard environment.

Future work involves designing a customized deep neural network architecture, which is optimized for detection of different orchard flowers and fruits.

## 7 Acknowledgements

This research was supported by the New Zealand Ministry for Business, Innovation and Employment (MBIE) on contract UOAX1414.

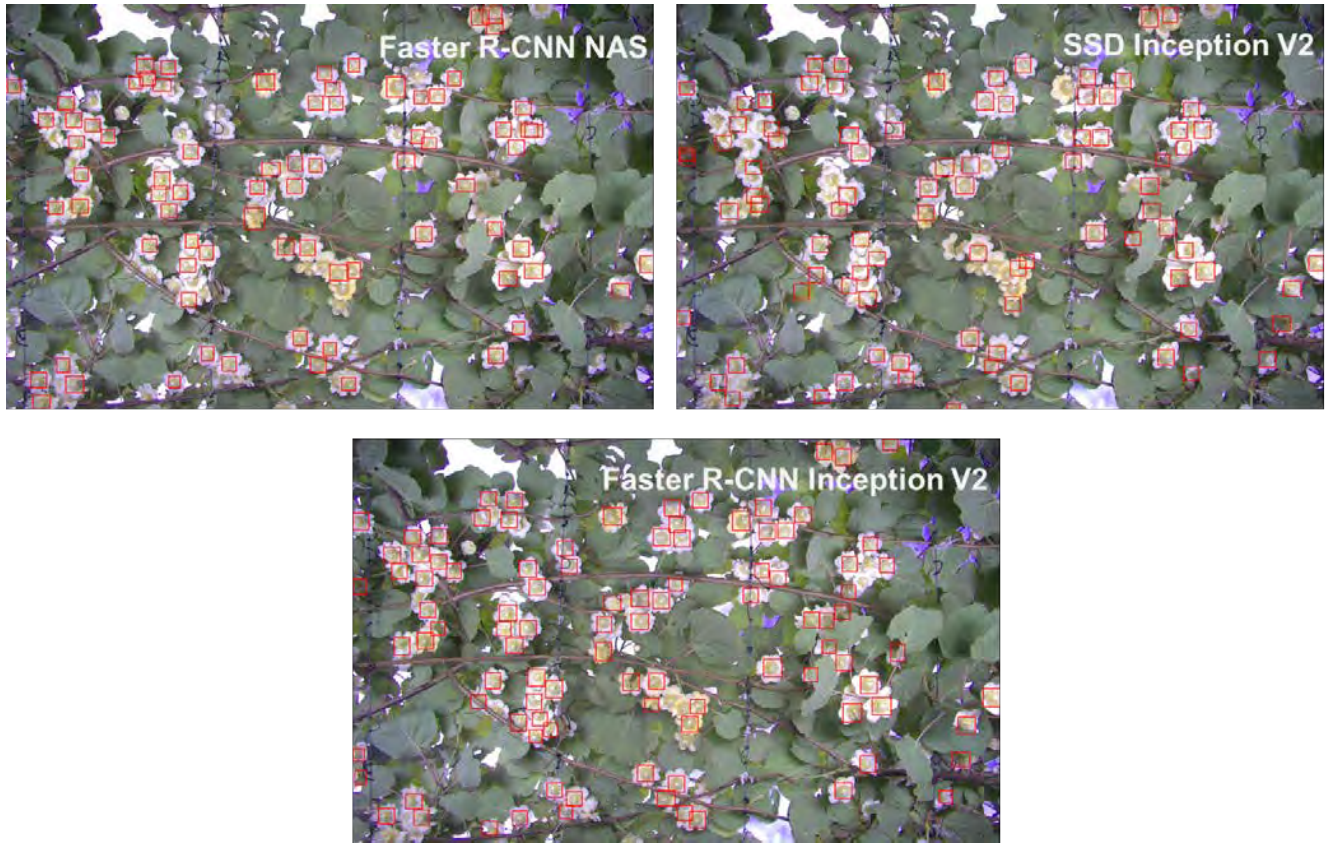

Fig. 9. Inference results using Faster R-CNN NAS, SSD Inception V2 and Faster R-CNN Inception V2 on a same image.